\documentclass[letterpaper, 10 pt, conference]{ieeeconf}  

\IEEEoverridecommandlockouts                              

\usepackage{graphics} 
\usepackage{epsfig} 
\usepackage{mathptmx} 
\usepackage{times} 
\usepackage{amsmath} 
\usepackage{amssymb}  

\title{\LARGE \bf
VR-GPT: Visual Language Model for Intelligent Virtual Reality Applications
}

\author{Mikhail Konenkov, Artem Lykov, Daria Trinitatova, Dzmitry Tsetserukou 
\thanks{The authors are with the Intelligent Space Robotics Laboratory, Center for Digital Engineering, Skolkovo Institute of Science and Technology (Skoltech), 121205 Moscow, Russian Federation. 
E-mail: {\tt\small \{mikhail.konenkov, artem.lykov, daria.trinitatova, d.tsetserukou\}@skoltech.ru\
}
}
}

\begin{document}

\maketitle
\thispagestyle{empty}
\pagestyle{empty}

\begin{abstract}

The advent of immersive Virtual Reality applications has transformed various domains, yet their integration with advanced artificial intelligence technologies like Visual Language Models remains underexplored. This study introduces a pioneering approach utilizing VLMs within VR environments to enhance user interaction and task efficiency. Leveraging the Unity engine and a custom-developed VLM, our system facilitates real-time, intuitive user interactions through natural language processing, without relying on visual text instructions. The incorporation of speech-to-text and text-to-speech technologies allows for seamless communication between the user and the VLM, enabling the system to guide users through complex tasks effectively. Preliminary experimental results indicate that utilizing VLMs not only reduces task completion times but also improves user comfort and task engagement compared to traditional VR interaction methods. 

\end{abstract}

\section{Introduction}

In recent years, Virtual Reality (VR) has evolved from an entertainment opportunity into a critical tool across multiple industries including education \cite{radianti2020systematic}, \cite{pellas2021immersive}, \cite{soliman2021application}, healthcare \cite{javaid2020virtual}, \cite{hilty2020review}, \cite{mirchi2020virtual}, and manufacturing \cite{doolani2020review}, \cite{malik2020virtual}, \cite{dallel2023digital}. This transformation has been fueled by significant advancements in computing power and VR technologies, which made the integration of VR efficient, profitable, and accessible.  With the Artificial Intelligence (AI) emerging and evolving by recent breakthroughs in models' capabilities to show great performance among various disciplines \cite{openaisora}, \cite{claude3}, the integration of this development area with VR has begun to show promising avenues for enhancing user experience and operational efficiency. Among other AI technologies, Vision Language Models (VLMs) is one of the most promising due to its ability to combine computer vision (CV) and natural language processing (NLP) capabilities. These models are designed to understand and generate text about images, bridging the gap between visual information and natural language descriptions. The state-of-the-art (SOTA) models have been proven to be accurate and efficient to be implemented in various fields, including robotics \cite{shah2023lm}, autonomous driving \cite{sima2023drivelm}, and healthcare \cite{moon2022multi}.

\begin{figure}[thpb]
    \centering
    \includegraphics[width=0.45\textwidth]{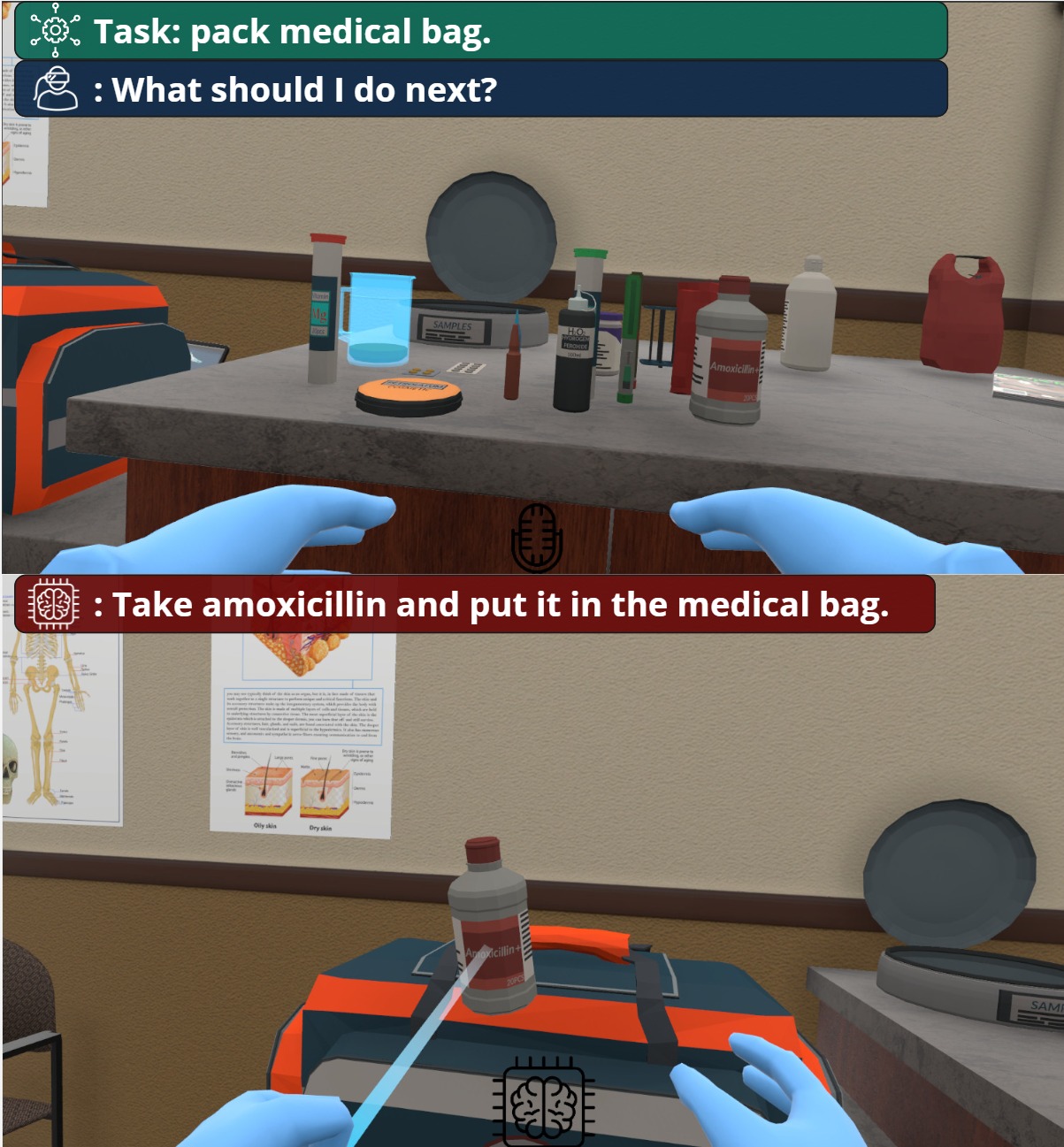}
    \caption{Application of VLM-VR integration system to the medical task.}
    \label{figurelabel}
    \vspace{-0.3cm}
\end{figure}

While VLMs have demonstrated their effectiveness in question answering, reasoning, and image answering on single image and on time-series images, its integration into interactive VR environments presents unique challenges. Even though existing research explores VLM applications across multiple domains, including challenging tasks of surgery assistance or road hazard prediction, the field of VR remains almost untouched by it, leaving a significant gap in this area of knowledge. This gap underscores a critical oversight, as the interactive nature of VR could significantly benefit from the nuanced understanding and response capabilities of VLMs. Challenges in this domain include the optimization of VLMs for real-time interaction, managing the computational demands within VR hardware constraints, and ensuring the models' responsiveness and accuracy in varied VR scenarios. Addressing these challenges is crucial for enabling more sophisticated human-computer interactions within these environments.

Current methods for interactions in VR usually include only pre-recorded audio or visual feedback. There are some existing methods which utilize large language models (LLMs) for generating responses, usually implemented as human or symbol avatar \cite{zhu2023free}. However, these methods are not aware of virtual space and cannot provide an answer based on the particular situation. As a result, there is a need for a more efficient method for VLM based instruction assistant technique that can provide information on the specific task in VR environment.

This study presents a novel approach that combines VLM and VR to assist the user on performing tasks in the virtual environment. The system utilizies Unity engine\cite{unity}, a real-time 3D development platform for building VR environments, and VLM to provide natural language interactions and task instructions to the user. The system is capable of advising the user theirs next action during the task in the form of audio response without visual text instructions, which is achieved with text-to-speech and speech-to-text (TTS and STT) technology. Besides, we collected a dataset, designed with the idea to improve model's capabilitities in world knowledge. Our approach offers the potential to use mixed reality (XR) as the useful tool in different tasks, including both VR and real-world applications.

\vspace{-0.1cm}
\section{Related Works} \label{sec:literature}

\subsection{VR}
While VR has not been taking advantage of the capabilities of the multimodal models, the research in this area is emerging in exploration of various 3D user interfaces (UI), multisensory and multimodal interactions, avatars, and many others. In the work of Zhu J. et al. \cite{zhu2023free} two types of avatars, human and symbolic, were designed to investigate the impact of visual representation of agents in mixed reality (MR). OpenAI's GPT-3 was used to facilitate the conversations. In \cite{roberts2022surreal}, OpenAI's Codex was used to dynamically transform ball and paddles in the pong game. The game also generates interactions based on the object's properties. The study \cite{mirchi2020virtual} explores the impact of AI on surgery and medicine training and highlights the benefits of real-time, automated feedback to users. The potential of combining AI and XR in application to learning also shown in the study by Rahui Divekar et al. \cite{divekar2022foreign}, where multimodal model was implemented as immersive agent for studying Chinese. The results demonstrated the statistically significant improvement both in students' language skills and in engagement.

In the field of VR interaction methods many works have developed robust systems which improve the user experience and results in the large spectre of tasks. In \cite{rodrigues2023amp} the authors propose two novel techniques for high-precision manipulation. Results indicate that these have significant advantages over best-practice techniques regarding task performance and user preference. In other work, Ponomareva et al. \cite{ponomareva2021grasplook} presented novel VR-based telemanipulation system, using R-CNN for object detection and pose estimation. The developed system has decreased task completion time and increased the users' satisfaction in the telemanipulation task.

\subsection{VLMs and Multimodal Models}

The interest in the multimodal models and its applications have been rapidly increasing since GPT-4 release \cite{achiam2023gpt}, and currently there are a lot of papers which explore the capabilities of such models in different scenarios. In \cite{bai2023qwen}, the multimodal model, Qwen-VL, was tested on various visual understanding tasks and outperformed other models on several benchmarks. The work of Chonghao Sima et al. \cite{sima2023drivelm} designed a novel framework for Graph Visual Question Answering, using Qwen-VL as part of the system. The approach has shown great results in complex driving scenarios and capable of predicting the movement on the road, safety, best driving strategy and path. The multimodal model was also used in \cite{shah2023lm}, where large pre-trained models were utilizied for robotic navigation. The system has demonstrated its ability to successfully navigate the robot through outdoor and suburban settings.

More advanced works explore the foundation models, which incorporate language, vision, and other type of data for completing sophisticated tasks in various domains. The work of Google Deepmind AutoRT \cite{ahn2024autort} leverages VLMs for scene analysis and LLMs for proposing diverse instructions to the fleet of 20 robots in unseen scenarios. Moreover, they collected 77k episodes of real-world teleoperation and autonomous robot policies and show, that such approach significantly increases the effectiveness of action input in the foundation model. In the work CognitiveDog \cite{lykov2024cognitivedog}, and in the following work CognitiveOS \cite{lykov2024cognitiveos} the authors applied VLM and LLM as modules of the multi-agent system for tackling intricate real-world tasks. In the more recent work there are nine different modules, including question answering, appropriate ethical behaviour analysis, object localization, and others, which increases the system's adaptability and performance.

\section{System Overview} \label{sec:sys}

\begin{figure}[thpb]
    \centering
    \vspace{0.2cm}
    \includegraphics[width=0.48\textwidth]{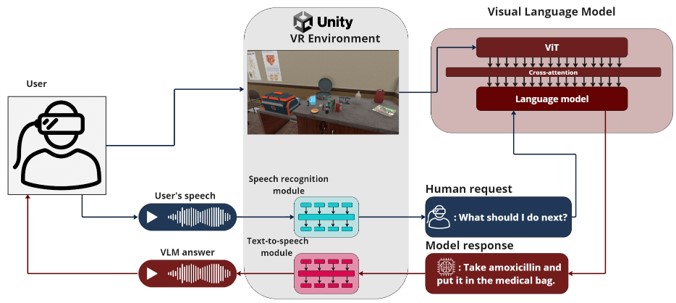}
    \caption{System Architecture.}
    \label{fig:arch}
    \vspace{-0.5cm}
\end{figure}

This section provides a comprehensive System Overview of our approach. The key feature is VLM integration with Unity engine, in order to facilitate complex tasks within VR environment. VLM acts a core of the system, operating on a server and handling requests from the Unity engine. We chose Qwen-VL \cite{bai2023qwen} with 7B parameters due to its outstanding abilities in visual question answering and reasoning. Using UnityWebHandler, we were able to process the dialogue between user and model fast, which made the experience more immersive and decreased the task completion time. The system architecture is shown in Fig. \ref{fig:arch}. The VLM runs on the server equipped with NVIDIA GeForce RTX 4090 24Gb GPU Memory, while the VR environment in Unity runs on the PC with NVIDIA GeForce GTX 1080 12Gb GPU Memory.

\subsection{User-VLM Interaction}

Since the instructions should not distract the user and be as natural as possible, we implemented the STT and TTS for messages exchange between user and VLM. The Jets model was applied for TTS. The Jets is based on the ESPnet \cite{watanabe2018espnet}, end-to-end speech processing toolkit, and inherits the software architecture from it. User inputs are captured through speech, converted to text using the Sentis plugin utilizing Whisper \cite{radford2023robust} — approach for robust speech recognition with large-scale weak supervision, developed by OpenAI. We utilized Whisper tiny model with 39M parameters, since it allowed to increase the performance without large quality loss. Since it requires the recorded audio to be 16kHz WAV format, we also implemented the resampling model to transform the recorded user's request. The processed requests are sent to the VLM, which interprets the user's intent and generates appropriate responses or actions. These responses are then converted into speech using Jets, providing auditory feedback to the user within the VR environment.

\subsection{Environment Interactions}

Interactions within the VR environment are facilitated by the XR Interaction Toolkit, which enables natural and intuitive user interactions such as grabbing and manipulating virtual objects. The user moves around with constant movement algorithm, which also makes the experience more organic. The buttons on the controller are attached to main system's functions, including grab mechanics, rapid turning, and recording the voice message, which is further transformed to text and sent to the VLM server. The Meta Quest 2 headset is utilized as the primary interface, offering high-resolution visuals and responsive tracking to immerse users fully in the VR tasks.

\subsection{VR Environments} \label{sec:env}

The system is demonstrated across two primary VR settings: a kitchen and a laboratory, in which the corresponding objects are placed on the working tables. In both environments, the interactions mechanisms, such as grab, are implemented. A critical component of the system is the task manager, which monitors and records the progress and completion of tasks and subtasks. Although the task manager does not interact directly with the VLM, it serves as an organizational backbone, ensuring that tasks are completed in a logical sequence and that the user is systematically guided through the task pipeline. In order to evaluate user success on completing the actions, event trigger system is used. Usually the task consists of several pick-and-place actions, and for each object involved in the action the collision is calculated. If the collusion occurs, the action is considered completed. This applies both to the order-based tasks and to tasks, which are not order-dependent. The timer for task completion speed is implemented, but not shown to the user to avoid its influence on the experience. 

\subsection{Dataset}\label{dataset}

 To finetune the VLM we collected the dataset composed of the images taken in two developed VR environments: kitchen and medical laboratory. The choice of environments is motivated by the need for our system to be able to perform well both for general and case-specific tasks.
 
 The dataset includes images of the VR setting from different scenarios. In each scenario at least 6 actions could be taken in order to complete the task. Every completed action was counted as another step in the task's progress. For each step, the images were taken from 4 different points: left, right, center and top. 
 
In order to cover more ways for the user to interact with the system, three different groups of text prompts were composed for each image of the collected dataset. In all three groups the initial sentence stated the goal (e. g. 'The task: collect all fruits in the wooden bowl'). In the first group the next sentence was imitating the user's request about next action (e. g. 'What is the next step?'), and the output was the action suggestion. In the second group, the user's request was about finding the object(e.g. 'Where is the apple?'), and the response consisted of the object's name and coordinates. In the third group, besides the main goal, the list of previous actions is transferred. That allows our system to be used in tasks where the strict order of the actions required.

 This system's design aims to streamline the VR experience, making it more accessible and productive for users by leveraging advanced VLM capabilities to understand and respond to natural language within immersive environments.

\section{Evaluation}

\subsection{Model's instructing ability}

\begin{figure}[ht]
 \centering
  \includegraphics[width=0.45\textwidth]{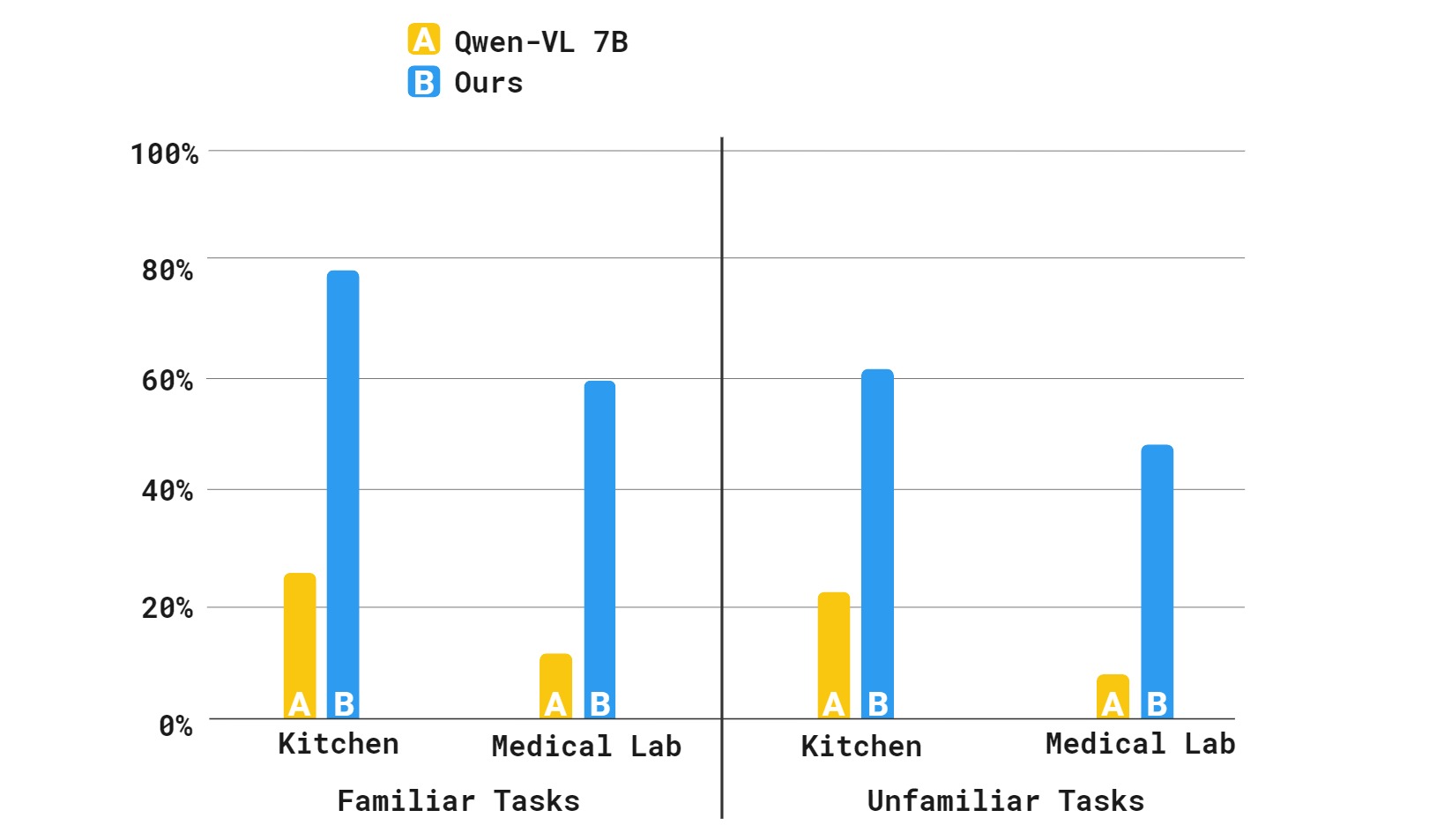}
  \caption{Performance comparison for two categories of tasks for different environments.}
  \label{fig:res1}
\end{figure}

In order to complete preliminary tests on the system's instructing abilities while operating in VR environment, the initial testing should be conducted after the model's finetuning. The structure of the dataset collected in developed VR environments was described in Sec. \ref{dataset}. The samples were split for training, validation and test set in 80-10-10\% accordingly. Moreover, to evaluate the model's ability to suggest actions for tasks not used in the training set, an additional test dataset was collected with two unique scenarios, one for each environment. For each step of the scenario, the list of correct next actions was compiled, and the model responses during the experiment were compared with these lists of actions. We took the not fine=tuned Qwen-VL model as the baseline, using the one-shot learning in the input.  The experimental results are depicted in Fig. \ref{fig:res1}.

 The results show that our system performed well both in familiar and unfamiliar scenarios, which proves its capabilities of environment analysis and instruction. The success rate for the familiar tasks category is 78\% and 60\% for the kitchen and medical laboratory environments respectively, while for the unfamiliar tasks the success rate is 62\% and 48\%. On average, our system outperformed the baseline by more than 3 times. We also note, that both the baseline and the fine-tuned model show greater results for tasks in the kitchen environment. We attribute this interesting result to two factors, which are the complexity of the task and the environment uniqueness. To solve narrowly focused tasks in specific environments, the system requires additional finetuning.
 
\subsection{User Study on VLM User Interaction}

The user study was conducted to compare our system with the standard approach for task instruction in VR environments. We took text-based instructions as the baseline approach, and developed two scenarios in the kitchen and medical lab environments. In both scenarios, the task consisted of 6 pick-and-place actions. The order of the interaction methods and environments was randomly chosen for every single experiment run. In addition, after completing the tasks with both interaction methods, the participants were asked to complete a 9-question survey  using a five-point Likert scale. 

\subsection{Participants}
The experiment involved 6 subjects (5 male and 1 female). The average participant age was 25.67 with a range 24-29. In total, one participant never were immersed in VR, three were using a VR once or several times for entertaining purposes, and two reported regular experience with VR.

\subsection{Experimental setup}

Experimental setup included two environments: kitchen and medical office. In the kitchen environment, three groups of the object were laying on the kitchen table, representing three types of food: vegetables, fruits, and deserts.  In the medical office, besides medical equipment, there were pill bottles, vitamins, and cremes.

The baseline approach was implemented using the task manager described in Sec. \ref{sec:env}. It was utilized as the dropdown list, consisting of the task and the list of actions needed to complete the task. This list could be opened or closed using the button on the controller, and was attached to the user's view in order to simplify the procedure and make the list easily reachable. When one of the actions was completed, the corresponding action text was crossed out of the list.

The VR-GPT approach provided no visual feedback for task completion. During the conversation with the VLM, only voice requests and responses were used. As mentioned in Sec. \ref{sec:sys}, the button on the controller was used to record the voice message, which then was resampled to the 16kHz WAV file and further transformed with speech recognition module. The model response was transformed to the audio using TTS module and played for the user after. To provide additional feedback, the audio sound was played when one of the actions was completed. The participant could request the system the next step or information on object's position in the environment.

Interaction between the objects was implemented with the event trigger method in Unity. When one of the action objects (one that is picked during the action sequence used in the scenario) collided with the target object, which represented the place, the action was considered completed. If the object was chosen wrong and was not used in any action, the feedback to the user was provided either with the text message or with the sound depending from the used method. When all of the actions were finished, the task was complete. 

\subsection{Experimental procedure}

All the participants first completed the training in the separate special environment. They learned to interact with the objects and use both of the approaches by pressing button on the controller. After the training course, the participants were immersed into one of the experiment rooms and were asked to complete the task. The task in the kitchen was to collect all the fruits to the wooden bowl. The task in the medical office was to collect the vitamins and pill bottles to the medical bag. The participants could complete the tasks in any order, using baseline approach for the one environment and VR-GPT for another. For each participant, we measured the execution time and the number of wrong actions.

\subsection{Experimental results}

\begin{table}[htbp]
\caption{Comparison of Time of Operation and Number of the Wrong Actions for 2 VR Interaction Approaches}
\begin{center}
\begin{tabular}{|c|c|c|c|c|p{1.3cm}|}
\hline
\textbf{Interaction Approach}&\multicolumn{4}{|c|}{\textbf{Task execution time,s}}&{\textbf{Number of wrong actions}} \\
\cline{2-5} 
  & min & max & mean & sd & \\
\hline
Baseline & 53.2 & 164.4 & 112.6 & 34.21 & 2.6 \\
\hline
VR-GPT (Ours) & 56.5 & 142.2 & 96.8 & 25.34 & 1.7 \\
\hline
\end{tabular}
\label{table:exp2_1}
\end{center}
\end{table}

After conducting the experiment with both approaches, we compared the task execution time and the number of wrong actions (\ref{table:exp2_1}. The experiment results showed that the system allows to lower the execution time by 13.7\%, while also reducing the number of errors during the task completion. We also note that the minimum amount of time required for operation is lower for the baseline than for our approach. This result can be explained by the task simplicity in the case of the kitchen environment, where the text-based instructions can be faster implemented than chat with the model. However, the maximum amount of time is lower by 20.8\% for VR-GPT approach, which shows the system's capability to be implemented for interaction during comprehensive tasks completion.

\section{Conclusion}

This study successfully demonstrates the substantial benefits of integrating VLM into VR environments. Our system, VR-GPT, has not only shown a significant decrease in task completion time but also improved overall user engagement and comfort. These improvements are critical in environments where intuitive and effective user interaction is important. By leveraging VLM capabilities, VR-GPT allows users to interact naturally and efficiently within immersive settings without the distractions of standard text-based interfaces.

Our experiments confirm that VR-GPT significantly outperforms traditional interaction methods, making it a promising solution for various VR applications. Particularly, the system's ability to adapt to both familiar and unfamiliar scenarios underscores its robustness and flexibility. The improved performance in tasks within the kitchen and medical laboratory VR settings suggests that VR-GPT can be effectively tailored to specific environments.

\addtolength{\textheight}{-12cm}   





\bibliographystyle{IEEEtran}
\bibliography{root}

\end{document}